\documentclass[11pt]{article}

\usepackage[accent=PrincetonOrange,runningtitle={Automatic Generation of High-Performance RL Environments}]{preprint}

\graphicspath{{figures/}}

\newcommand{\seth}[1]{}
\newcommand{\rahul}[1]{}

\title{Automatic Generation of High-Performance RL Environments}

\author{Seth~Karten$^{1}$ \and Rahul~Dev~Appapogu$^{2}$ \and Chi~Jin$^{1}$\\[0.3em]
  \small $^{1}$Princeton University \quad $^{2}$Independent Researcher}

\date{\vspace{0pt}}

\correspondence{sethkarten@princeton.edu}

\begin{document}

\thispagestyle{empty}
\maketitle
\pagestyle{fancy}

\begin{abstract}
Translating complex reinforcement learning (RL) environments into high-performance implementations has traditionally required months of specialized engineering. We present a closed-loop methodology that produces equivalent high-performance environments for minimal compute cost. Our method uses a generic prompt template, hierarchical verification  (property, interaction, and rollout tests), iterative repair, and cross-backend policy transfer to verify no sim-to-sim gap. We demonstrate three distinct workflows across five environments: (1) Direct translation (no prior performance implementation exists) from Game Boy emulator PyBoy to our EmuRust (via Rust IPC) and from Pokemon Showdown to our PokeJAX (via JAX); (2) Translation verified against existing performance implementations via throughput parity with Puffer Pong, MJX and Brax at matched GPU batch sizes; and (3) New environment creation: TCGJax, the first Pokemon TCG Pocket environment, created from a web-extracted specification. At 200M parameters, the environment overhead drops below 4\% of training time. Our closed-loop methodology confirms equivalence for all five environments. TCGJax, synthesized from a private reference absent from public repositories, serves as a contamination control for agent pretraining data concerns.
\end{abstract}

\etocdepthtag.toc{mainpaper}
\section{Introduction}
\label{sec:intro}

In typical reinforcement learning (RL) training, environment simulation consumes 50--90\% of wall-clock time~\citep{lu2022discovered, suarez2024pufferlib}. For complex simulators, such as Pokemon Showdown~\citep{zarel2011pokemonshowdown,karten2025pok,karten2026pokeagent} at 100K+ lines of TypeScript, or cycle-accurate Game Boy emulators in C, this overhead is even more severe. Foundation RL architectures~\citep{reed2022generalist, grigsby2024amago} that train across many environments amplify the cost of slow simulation, motivating scalable methods for producing performance environments.

The RL community has responded with award-winning hand-optimized rewrites: Brax~\citep{freeman2021brax}, Gymnax~\citep{lange2022gymnax}, Pgx~\citep{koyamada2023pgx}, JaxMARL~\citep{rutherford2024jaxmarl}, Craftax~\citep{matthews2024craftax}, and PureJaxRL~\citep{lu2022discovered}. Each required labor-intensive specialized engineering for a single domain. A method for producing performance environments cheaply and routinely, as a standard step in the RL workflow, would complement existing libraries.

\begin{figure}[t]
\centering
\includegraphics[width=\textwidth]{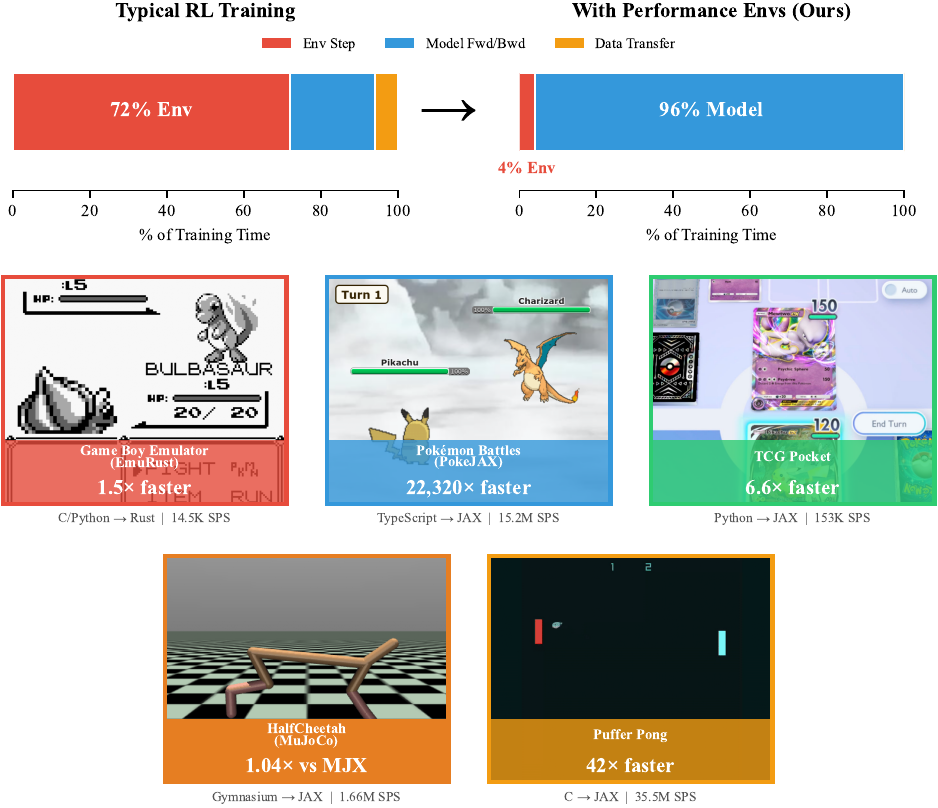}
\caption{\textbf{Performance environments eliminate the environment bottleneck.} (Top)~Our methodology shifts training from environment-bound to model-bound. (Bottom)~Five case studies, grouped by result type. (1)~\emph{Direct translation} (no prior performance implementation exists): EmuRust (Game Boy emulator via Rust IPC); PokeJAX (Pokemon Showdown via JAX). (2)~\emph{Translation verified against existing performance implementations}: throughput parity with Puffer Pong, MJX, and Brax at matched GPU batch sizes. (3)~\emph{New environment creation}: TCGJax, the first Pokemon TCG Pocket environment, created from a web-extracted specification.}
\label{fig:hero}
\end{figure}

Coding agents have demonstrated strong results on code generation and translation tasks, from competition-level programming~\citep{li2022competition} to resolving real-world software issues~\citep{jimenez2024swe} and large-scale code migration~\citep{ziftci2025migrating}. These capabilities extend naturally to RL environment translation: a single generic prompt (Appendix~\ref{app:prompts}) can produce a full environment rewrite for minimal compute cost. However, for RL environments, generation alone is insufficient. Silent errors in game mechanics or physics accumulate over thousands of timesteps and corrupt training signals. The critical gap is verification: without structured feedback confirming functional equivalence, the agent has no signal to iterate toward a correct translation.

We contribute: (1)~a closed-loop translation methodology that pairs hierarchical verification (property, interaction, and rollout tests) with cross-backend policy transfer, providing the structured error signal that enables agents to converge on equivalent high-performance environments; and (2)~empirical validation across three settings: direct translation where no prior performance implementation exists (EmuRust, PokeJAX), translation verified against existing optimized implementations (Pong, HalfCheetah), and new environment creation from a web-extracted specification (TCGJax). We validate our methodology with two additional coding agents on representative environments (Table~\ref{tab:multi_agent}), and provide sufficient detail, including representative prompts, verification methodology, and complete results, to reproduce the translations by providing a coding agent with a reference environment codebase, our methodology, and iterating until verification is complete.

\section{Related Work}
\label{sec:related}

\paragraph{Hardware-accelerated environments.}
A growing body of work manually reimplements RL environments in JAX or on GPU. Brax~\citep{freeman2021brax} rewrites rigid-body physics; MJX~\citep{todorov2012mujoco} ports MuJoCo to XLA; Gymnax~\citep{lange2022gymnax} reimplements classic control; Pgx~\citep{koyamada2023pgx} covers board games; JaxMARL~\citep{rutherford2024jaxmarl} provides multi-agent environments; Craftax~\citep{matthews2024craftax} reimplements Crafter ($250\times$ speedup); and PureJaxRL~\citep{lu2022discovered} demonstrated $4k\times$ speedup from JAX. Each required significant specialized engineering for a single domain. Our methodology produces parallelizable environments using a generic prompt template, enabling ease of creation for high performance environments from original implementations.

\paragraph{High-throughput RL systems.}
Gymnasium~\citep{towers2026gymnasium} standardizes the environment interface used by most RL libraries. EnvPool~\citep{weng2022envpool} achieves high throughput via C++ async batching; PufferLib~\citep{suarez2024pufferlib} provides a unified interface for C environments; Sample Factory~\citep{petrenko2020sample} maximizes GPU utilization. Our work is complementary: reducing per-step time lets these systems fully exploit their parallelism. We show our method is able to attain parity performance with these existing systems.

\paragraph{Sim-to-real transfer.}
The sim-to-real gap~\citep{tobin2017domain, zhao2020sim, da2025survey} measures how well policies trained in simulation transfer to physical systems, where differences in physics, rendering, and sensor noise degrade performance. Domain randomization and system identification are standard approaches to closing this gap. We address the analogous problem between two simulators: rather than building robustness to unmodeled dynamics, we verify exact behavioral equivalence between the reference and translated environment by training policies and transferring them across backends.

\paragraph{LLM-assisted code generation.}
Neural code translation~\citep{roziere2020unsupervised}, AlphaCode~\citep{li2022competition}, and SWE-bench~\citep{jimenez2024swe} address function- or API-level tasks. \citet{ziftci2025migrating} report ${\sim}50\%$ effort reduction from LLM-assisted migration at Google. Eureka~\citep{ma2024eureka} and Text2Reward~\citep{xie2024text2reward} use LLMs to generate reward functions. Our setting differs: we translate full RL environments where silent errors in game mechanics or physics accumulate over thousands of timesteps and corrupt training signals. Hierarchical verification closes this gap by localizing errors at the component level before they propagate to rollouts.

\section{Methodology}
\label{sec:method}

\begin{figure}[t]
\centering
\includegraphics[width=\textwidth]{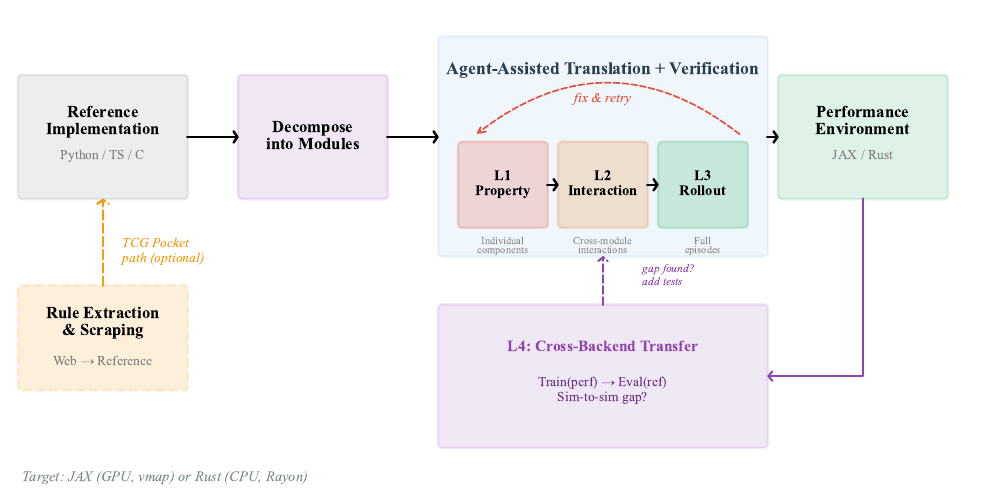}
\caption{\textbf{Translation and verification pipeline.} A reference environment is decomposed into modules, translated by a coding agent, and verified through four levels of increasing scope. Failures at any level trigger targeted repair and re-verification; Level~4 cross-backend policy transfer closes the outer loop.}
\label{fig:pipeline}
\end{figure}

We translate reference RL environments into high-performance equivalents via coding agents guided by hierarchical verification and sim-to-sim gap detection after training. Figure~\ref{fig:pipeline} summarizes the pipeline.

\subsection{Problem Statement}
\label{sec:method:problem}

Given a reference environment $E_{\text{ref}}$ in source language $L_{\text{src}}$, we produce a high-performance environment $E_{\text{perf}}$ in target language $L_{\text{tgt}}$ that is behaviorally equivalent: for any seed and action sequence, both environments produce identical observations, rewards, and termination signals at every timestep. For continuous-valued environments (e.g., physics simulations), we relax this to $\epsilon$-equivalence, where per-step outputs agree within a per-component $L_\infty$ tolerance $\epsilon$ (see \S\ref{sec:exp:effort} for environment-specific tolerances). The central question is whether any remaining differences affect training outcomes. Analogous to the sim-to-real gap in robotics~\citep{tobin2017domain, zhao2020sim}, we refer to this as the \textbf{sim-to-sim gap}: the difference in policy quality that arises from training in $E_{\text{perf}}$ versus $E_{\text{ref}}$.

To verify there is no sim-to-sim gap, we use cross-backend policy transfer: a policy trained in $E_{\text{perf}}$ is evaluated in $E_{\text{ref}}$, and vice versa. Both environments use identical hyperparameters and training algorithms, isolating the environment as the only variable. If reward is statistically indistinguishable across backends, there is no gap. We verify this empirically over 100 episodes with diverse random seeds rather than through formal proof. We additionally require that $E_{\text{perf}}$ achieves sufficient throughput to shift the training bottleneck away from environment simulation.

We select between JAX and Rust based on environment structure: JAX suits environments with small state memory that benefit from parallel execution on GPU; Rust suits sequential or memory-intensive environments. See Appendix~\ref{app:details} for detailed selection criteria.

\subsection{Agent Translation and Verification}
\label{sec:method:verification}

Exhaustive rollout comparison is intractable: the space of possible state and action sequences is too large to test completely. Instead, translation and verification proceed together through four levels of increasing scope, where cross-backend policy transfer tests equivalence under the state distribution that matters for training. Failures at any level trigger code correction before advancing.

\textbf{Level~1: Component generation (L1).} The agent translates individual components from $E_{\text{ref}}$ source code into $E_{\text{perf}}$, verifying each in isolation before proceeding.
\textbf{Level~2: Interaction tests (L2).} Tests verify functions and interactions between components, checking that composed modules produce correct outputs. Failures trigger repair while preserving L1 correctness.
\textbf{Level~3: Rollout comparison (L3).} Full episodes are executed in both environments under matched seeds and identical action sequences, comparing observations, rewards, and terminations at every timestep. Discrepancies trigger root-cause analysis with new targeted L1/L2 tests.
\textbf{Level~4: Cross-backend policy transfer (L4).} A policy trained in $E_{\text{perf}}$ is evaluated in $E_{\text{ref}}$ (and vice versa), confirming no sim-to-sim gap under learned policy behavior. A detected gap feeds back into L1--L3 until it closes.

The iterative cycle is required to converge to a correct translation. All translations used Gemini~3 Flash Preview, invoked via the Gemini CLI in non-interactive mode (\texttt{gemini --yolo}); however, the methodology is agent-agnostic. The agent receives module source code, target language specification, and test requirements in a single prompt (see \S\ref{sec:exp:effort} for measured costs). Human involvement is limited to writing translation prompts (what is the reference environment?) and designing verification test structures (what is the verification signal?). Appendix~\ref{app:prompts} provides a representative prompt, and Appendix~\ref{app:optimization} provides backend-specific optimization checklists. Algorithm~\ref{alg:translate} in Appendix~\ref{app:algorithm} formalizes this process.

\section{Experiments}
\label{sec:exp}
We evaluate four hypotheses across five environments spanning discrete games, continuous physics, hardware emulation, and multi-agent systems (Table~\ref{tab:env_overview}):
\textbf{H1:} Agent-generated environments are equivalent to the reference, with no sim-to-sim gap (\S\ref{sec:exp:policy}).
\textbf{H2:} Agent-generated environments achieve sufficient throughput to shift training from environment-bound to model-bound (\S\ref{sec:exp:throughput}, \S\ref{sec:exp:breakdown}).
\textbf{H3:} Hierarchical verification is necessary for convergence (\S\ref{sec:exp:effort}).
\textbf{H4:} The methodology generalizes across diverse environment types and translation settings (\S\ref{sec:exp:throughput}\,--\,\ref{sec:exp:effort}).

All benchmarks use 1$\times$ RTX 5090. Training curves use $N{=}10$ seeds with matched PPO~\citep{schulman2017proximal} hyperparameters. Additional details in Appendix~\ref{app:exp_details}.
\begin{table}[t]
\centering
\caption{\textbf{Environment overview.} $^{\star}$Private reference (contamination control).}
\label{tab:env_overview}
\footnotesize
\begin{tabular}{@{}lllrrl@{}}
\toprule
\textbf{Env} & \textbf{Source} & \textbf{Target} & \textbf{Src LoC} & \textbf{Tgt LoC} & \textbf{Key Challenge} \\
\midrule
EmuRust & C/Python & Rust+PyO3 & ${\sim}$26K & 2.5k & Cycle-accurate emulation \\
PokeJAX & TypeScript & JAX & ${\sim}$100K & 55k & state complexity. \\
HalfCheetah & MuJoCo & JAX & 245 & 1.2k & Articulated body + contact \\
TCGJax & Web rules$^{\star}$ & Py$\to$JAX & 29k & 4.2k & Rule extraction from web \\
Pong & C (PufferLib) & Rust+JAX & 225 & 235/318 & Already-optimized baseline \\
\bottomrule
\end{tabular}
\end{table}

These five environments were chosen to span the three translation settings and cover a range of complexity, state types, and compute backends. EmuRust translates a cycle-accurate Game Boy emulator (26K lines of C) into Rust, requiring exact hardware implementations across CPU, memory, and graphics subsystems. PokeJAX is the largest translation (100K+ lines of TypeScript, 55K lines of JAX), flattening a client-server architecture into topologically-sorted functions that enable 1,370 move effects via JAX's \texttt{lax.switch}. HalfCheetah requires reimplementing articulated-body dynamics with ground contact in pure JAX from the MuJoCo source code (and specifying the target MJX physics simulator), where small errors in physics propagate across simulation substeps. TCGJax tests specification-to-implementation translation: rules were extracted from web sources into a Python reference, then translated to JAX. Its private reference also serves as a contamination control for agent pretraining data. Pong is intentionally simple, testing whether our methodology produces competitive results against an already-optimized C baseline. Per-environment architectural details are in Appendix~\ref{app:env_details}.

\subsection{Throughput Results}
\label{sec:exp:throughput}

We evaluate \textbf{H2} (throughput) and \textbf{H4} (generalization) by measuring steps per second across all five environments (Table~\ref{tab:throughput_all}).

\begin{table}[t]
\centering
\caption{\textbf{Throughput comparison.} Mean $\pm$ std from $N{=}5$ runs (CVs ${<}3\%$); ${\sim}$2M models; JAX excludes JIT warm-up.}
\label{tab:throughput_all}
\small
\begin{tabular}{@{}llrrr@{}}
\toprule
\textbf{Environment} & \textbf{Benchmark} & \textbf{Reference (SPS)} & \textbf{Performance (SPS)} & \textbf{Speedup} \\
\midrule
\multicolumn{5}{@{}l}{\textit{Direct translation into newly performant environments (no prior performance implementation)}} \\
\midrule
\multirow{2}{*}{EmuRust}
  & Random action    & 167K (PyBoy, 32p)    & $239 {\pm} 6$K (Rust, 64e)    & $1.4\times$ \\
  & PPO training     & 9.9K (PyBoy, 32p)    & $14.5 {\pm} 0.4$K (Rust, 128e)  & $1.5\times$ \\
\midrule
\multirow{2}{*}{PokeJAX}
  & Random action    & 21K (Showdown, 1p)   & $500 {\pm} 9$M (JAX, 65Kb)    & $23k\times$ \\
  & PPO training     & 681 (Showdown)       & $15.2 {\pm} 0.2$M (JAX)         & $22k\times$ \\
\midrule
\multicolumn{5}{@{}l}{\textit{Translation verified against existing performance implementations}} \\
\midrule
\multirow{2}{*}{Puffer Pong}
  & GRU Rollout (2M) & $4.5 {\pm} 0.008$M (C, CPU)  & $140 {\pm} 1.5$M (JAX, GPU)     & $31\times$ \\
  & GRU PPO (2M)     & $854 {\pm} 4$K (C, CPU)      & $35.5 {\pm} 0.3$M (JAX, GPU)    & $42\times$ \\
\midrule
\multirow{3}{*}{HalfCheetah JAX}
  & vs Gymnasium     & 45K (1 proc)         & $1.66$M (JAX, 32Kb)   & $37\times$ \\
  & vs Brax          & 160K (Brax, 4Kb)     & $798$K (JAX, 4Kb)    & $5.0\times$ \\
  & vs MJX (Google)  & $1.6$M (MJX, 32Kb)    & $1.66$M (JAX, 32Kb)  & $1.04\times$ \\
\midrule
\multicolumn{5}{@{}l}{\textit{New environment creation (no prior trainable RL env)}} \\
\midrule
\multirow{2}{*}{TCGJax}
  & Random action    & 140K (Python, 16p)   & $717 {\pm} 0.6$K (JAX, 16Kb)    & $5.1\times$ \\
  & PPO training     & 23K (Python, 16p)    & $153 {\pm} 5$K (JAX, 4Kb)     & $6.6\times$ \\
\bottomrule
\end{tabular}

\vspace{1mm}
{\footnotesize p = processes, e = env instances, b/Kb = JAX batch (thousands). EmuRust comparison at matched 32 CPU cores.}
\end{table}

Results span three settings, supporting \textbf{H4}. \emph{Direct translations} (EmuRust, PokeJAX) produce performant versions where none existed; PokeJAX's $23k\times$ speedup reflects a shift from sequential CPU server to GPU-parallel pure functions, enabling convergent training previously impractical at 681 SPS. \emph{Verified translations} (Pong, HalfCheetah) achieve speedups over already-optimized baselines: Pong achieves $42\times$ end-to-end PPO via JAX training; HalfCheetah reaches throughput parity with Google's MJX ($1.04\times$), demonstrating that agent-generated code matches hand-optimized engines. \emph{New environment creation} (TCGJax) translates a web-extracted specification into a trainable JAX environment. These results support \textbf{H2}: all performance implementations achieve throughput sufficient to remove the environment as the training bottleneck. Per-environment details are in Appendix~\ref{app:env_details}.

\subsection{Training Time Breakdown}
\label{sec:exp:breakdown}

\begin{figure}[t]
\centering
\includegraphics[width=0.9\textwidth]{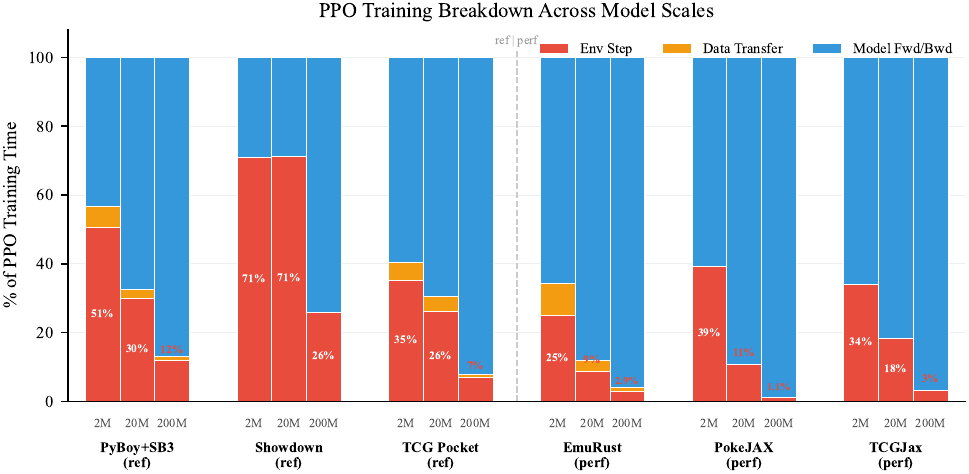}
\caption{\textbf{PPO training time breakdown across model scales.} Three bars per implementation show 2M, 20M, 200M parameter models. Performance implementations drop to ${\leq}4\%$ env overhead at 200M. All on 1$\times$ RTX 5090.}

\label{fig:breakdown}
\end{figure}

To further evaluate \textbf{H2}, Figure~\ref{fig:breakdown} profiles PPO iteration time across model scales (2M, 20M, 200M parameters). At 200M, all single-agent performance implementations contribute ${\leq}4\%$ of training time (down from 50--90\% for references), confirming that training has shifted from environment-bound to model-bound.

\subsection{Policy Equivalence}
\label{sec:exp:policy}

\begin{figure}[t]
\centering
\includegraphics[width=\textwidth]{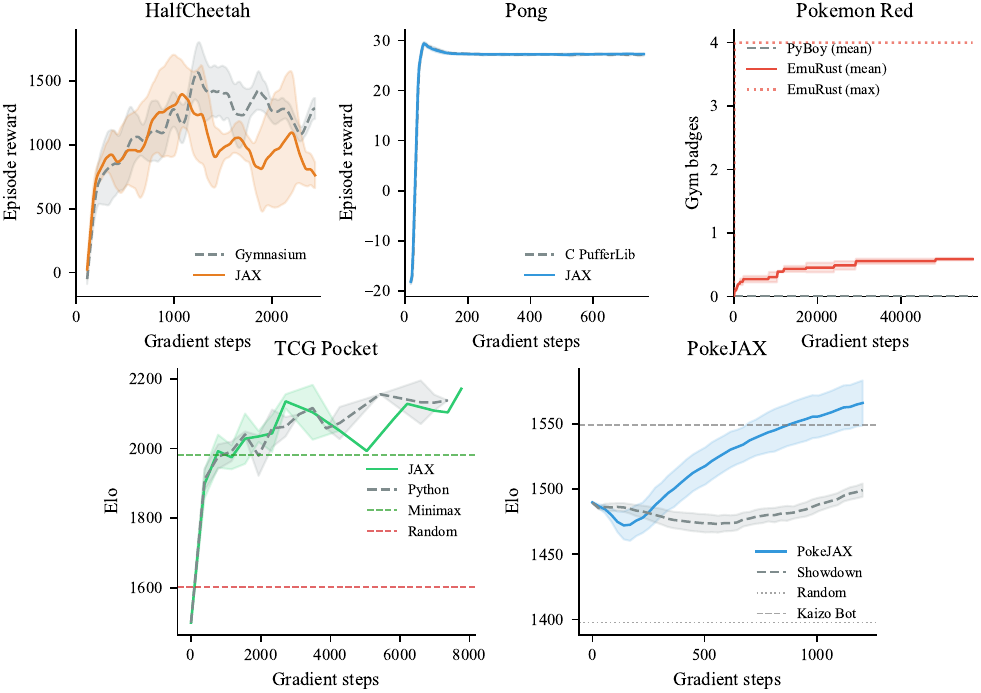}
\caption{\textbf{Policy equivalence.} Pong (10 seeds), HalfCheetah (10 seeds), EmuRust (10 seeds): matched reward curves across backends. TCGJax and PokeJAX: matched Elo curves (JAX vs reference). All five environments achieve L4 cross-backend transfer (Table~\ref{tab:cross_transfer}).}
\label{fig:training_curves}
\end{figure}

We evaluate \textbf{H1} (equivalence) through two complementary tests. First, all five environments pass L3 rollout comparison (100 episodes, matched RNG seeds, step-level output comparison; exact for discrete envs, $\epsilon{=}10^{-3}$ for HalfCheetah). Figure~\ref{fig:training_curves} shows matched training curves: Pong (10 seeds), HalfCheetah (10 seeds), and EmuRust (10 seeds) confirm consistent learning dynamics across backends.

\paragraph{Cross-backend policy transfer (L4).} Second, Table~\ref{tab:cross_transfer} evaluates policies trained in one backend on both backends (10 seeds each). We use the TOST (Two One-Sided Tests) equivalence procedure~\citep{schuirmann1987comparison} with environment-specific margins $\Delta$ (caption of Table~\ref{tab:cross_transfer}); a significant TOST result ($p{<}0.05$) confirms that the two backends produce equivalent performance within $\pm\Delta$. All five environments pass, supporting \textbf{H1}: Pong shows zero sim-to-sim gap, HalfCheetah confirms equivalent transfer despite high variance, and EmuRust-trained Pokemon Red policies transfer to PyBoy with near-identical reward. PokeJAX achieves \emph{exact} transfer: win rates are identical across backends. TCGJax likewise confirms equivalence in both directions.

\begin{table}[t]
\centering
\caption{\textbf{Cross-backend policy transfer.} Values are mean $\pm$ std over 10 seeds. Equivalence confirmed via TOST ($\alpha{=}0.05$) with environment-specific margins ($\Delta$): Pong $\Delta{=}1.0$, HalfCheetah $\Delta{=}100$, EmuRust $\Delta{=}0.5$, PokeJAX $\Delta{=}0.02$, TCGJax $\Delta{=}0.05$. PokeJAX and TCGJax report win rate against a heuristic bot; others report return.}
\label{tab:cross_transfer}
\small
\begin{tabular}{@{}llccc@{}}
\toprule
\textbf{Environment} & \textbf{Train Backend} & \textbf{Eval (Perf)} & \textbf{Eval (Ref)} & \textbf{Equiv.} \\
\midrule
\multirow{2}{*}{Puffer Pong} & C (ref) & $28.01 \pm 0.28$ & $28.04 \pm 0.29$ & $\checkmark$ \\
                              & JAX (perf) & $28.23 \pm 0.18$ & $28.22 \pm 0.20$ & $\checkmark$ \\
\midrule
\multirow{2}{*}{HalfCheetah} & MJX (ref) & $1398 \pm 497$ & $1389 \pm 511$ & $\checkmark$ \\
                              & JAX (perf) & $1026 \pm 636$ & $1133 \pm 562$ & $\checkmark$ \\
\midrule
\multirow{2}{*}{EmuRust (Red)} & PyBoy (ref) & $12.01 \pm 0.12$ & $11.99 \pm 0.15$ & $\checkmark$ \\
                                & Rust (perf) & $12.06 \pm 0.00$ & $12.06 \pm 0.01$ & $\checkmark$ \\
\midrule
\multirow{2}{*}{PokeJAX} & Showdown (ref) & $0.313 \pm 0.007$ & $0.313 \pm 0.007$ & $\checkmark$ \\
                          & JAX (perf) & $0.406 \pm 0.003$ & $0.406 \pm 0.003$ & $\checkmark$ \\
\midrule
\multirow{2}{*}{TCGJax} & Python (ref) & $0.575 \pm 0.054$ & $0.543 \pm 0.045$ & $\checkmark$ \\
                          & JAX (perf) & $0.583 \pm 0.062$ & $0.558 \pm 0.042$ & $\checkmark$ \\
\bottomrule
\end{tabular}
\end{table}

\subsection{Translation Effort and Verification}
\label{sec:exp:effort}

We evaluate \textbf{H3} (necessity of hierarchical verification) alongside translation cost. Table~\ref{tab:effort} summarizes cost across all five environments. All environment code is agent-generated; no lines were written by hand. Costs include all translation iterations (e.g., HalfCheetah required four solver revisions; EmuRust required three fix cycles).

\begin{table}[t]
\centering
\caption{\textbf{Translation cost.} Costs include all iterations; base rates from Gemini~3 Flash Preview logs for environments requiring multiple revision cycles.}
\label{tab:effort}
\small
\begin{tabular}{@{}lrrrrr@{}}
\toprule
\textbf{Metric} & \textbf{EmuRust} & \textbf{PokeJAX} & \textbf{HalfCheetah} & \textbf{TCG} & \textbf{Pong} \\
\midrule
Target LoC          & 2k    & 55k   & 1k & 4.2k & 235/318 \\
Modules             & 5          & 30     & 5 & 11 & 1 \\
Total tests         & 52         & 2k    & 69 & 50 & 12 \\
Agent cost          & \$0.43 & \$6 & \$3.26 & \$4.98 & \$0.05 \\
Agent iterations    & 72         & 63    & 20 & 51 & 13 \\
\bottomrule
\end{tabular}

\vspace{1mm}
\end{table}

Verification scope is summarized in Table~\ref{tab:verification} (Appendix~\ref{app:verification_summary}); all five environments pass all levels. Supporting \textbf{H3}, hierarchical verification is necessary for complex environments. On HalfCheetah, L3-only verification failed to converge after 42 iterations: the agent could not isolate dynamics bugs (e.g., a Coriolis force sign error) from end-to-end rollout failures, cycling through vectorization rewrites and stability patches instead. With the full hierarchy, L1 property tests caught these errors immediately (mass matrix symmetry, bias force magnitude bounds), and the translation converged in 5 iterations (Appendix~\ref{sec:exp:ablation_detail}). The failure mode is general: any environment with interacting physics subsystems produces rollout errors that are ambiguous without component-level diagnostics. Supporting \textbf{H4}, the methodology is agent-agnostic: re-translating Pong with Claude Sonnet~4.6 and HalfCheetah with Claude Opus~4.6 produces functionally correct translations using identical prompts (Table~\ref{tab:multi_agent} in Appendix~\ref{app:multi_agent}).

\section{Conclusion}
\label{sec:conclusion}

We presented a closed-loop methodology for translating reference RL environments into equivalent high-performance implementations using coding agents guided by hierarchical verification. Our two contributions are: (1)~the methodology itself, pairing four levels of verification (component generation, interaction tests, rollout comparison, and cross-backend policy transfer) with iterative repair to close the sim-to-sim gap; and (2)~empirical validation across three settings: direct translation where no prior performance implementation exists (EmuRust, PokeJAX), translation verified against existing optimized implementations (Puffer Pong, HalfCheetah), and new environment creation from a web-extracted specification (TCGJax).

Across five environments, we confirmed all four hypotheses. Agent-generated environments show no sim-to-sim gap (H1), verified by cross-backend policy transfer with TOST equivalence for all five environments. Performance implementations achieve throughput sufficient to shift training from environment-bound to model-bound (H2), with end-to-end PPO speedups from $1.5\times$ to $42\times$ and throughput parity with Google's MJX. Hierarchical verification is necessary for convergence on complex environments (H3), as shown by the HalfCheetah ablation where L3-only verification failed after 42 iterations. The methodology generalizes across environment types and coding agents (H4).

The methodology is most effective for environments with reproducible transitions, clear module boundaries, and fixed-size state representations. Environments with non-deterministic external dependencies (network calls, hardware-in-the-loop) or unbounded dynamic allocation may require additional engineering. Our equivalence verification depends on the quality of the trained policy: a stronger policy reaches states that random rollouts or weak policies never visit, potentially revealing gaps that current L4 evaluation does not catch. As training improves, re-running L4 with better policies provides a stronger verification signal. Additionally, not all environments are guaranteed to be faster after translation, particularly those already hand-optimized by skilled engineers. Our HalfCheetah result ($1.04\times$ vs MJX) illustrates this: against a mature, hand-optimized engine, the methodology achieves parity rather than speedup. The approach is best applied to unoptimized or new environments where no performance implementation exists.

The methodology decouples environment complexity from training cost: researchers can produce performance versions of the environments they need, rather than being limited to existing JAX ports. Re-translating when a reference updates costs under \$1, with the test suite serving as a regression guard. As coding agents improve and per-token costs fall, fast verified simulation becomes a default step in the RL workflow rather than a bottleneck requiring months of specialized engineering.


\bibliographystyle{abbrvnat}
\bibliography{main}

\clearpage
\appendix
\etocdepthtag.toc{appendixpart}
\appendix
\section{Supplementary Details}
\label{app:details}

This appendix provides additional tables, figures, and detailed descriptions that support the main text.

\subsection{Per-Environment Details}
\label{app:env_details}

The following paragraphs provide detailed descriptions for each environment, complementing the summary in \S\ref{sec:exp:throughput}.

\paragraph{EmuRust (C/Python $\to$ Rust).} The Game Boy emulator decomposes into five modules (CPU, memory, PPU, core, bindings). Both reference and translation run on CPU: PyBoy~\citep{baekalfen2018pyboy} uses Python multiprocessing (one process per instance), while EmuRust uses Rayon's work-stealing thread pool within a single process. The $1.5\times$ comparison is at matched CPU resources: both backends use the same 32 cores, but PyBoy saturates at 32 processes (one per core) while EmuRust packs 128 environments into a single process via Rayon's shared-memory thread pool, achieving higher per-core utilization through cooperative scheduling with zero IPC overhead.

\paragraph{PokeJAX (TypeScript $\to$ JAX).} PokeJAX is the first GPU-parallel Pokemon battle simulator. The standard tool for RL researchers was Pokemon Showdown~\citep{zarel2011pokemonshowdown}, a TypeScript server designed for human online play that has since become the primary testbed for competitive Pokemon AI, though not originally designed for RL training. Translating it (100K+ lines) required server/client flattening, fixed-size state arrays, and branch-parallel dispatch via \texttt{jax.lax.switch}. The full 55k-line translation is complete across ${\sim}$30 modules; only minor rule-edge-case modules are excluded. The ${\sim}$\$6 cost in Table~\ref{tab:effort} is extrapolated from a 5-module subset for which session-level cost logs were available. The reference baseline (21K SPS) reflects a single-threaded server not designed for throughput; running multiple instances via PokeEnv yields only 681 SPS due to WebSocket overhead. The $23k\times$ is an \emph{enabling number}: without this translation, training a Pokemon battling agent is impractical. The speedup decomposes into JAX compilation + GPU batching at 1K instances ($560\times$) and batch scaling from 1K to 65K ($42.5\times$), reflecting an architectural change (sequential CPU server to GPU-parallel pure functions), not per-instruction optimization. The 1,370 move functions dispatched via \texttt{lax.switch} produce a large XLA HLO graph, reflected in the 45\,s JIT time; every step pays the cost of all 1,370 move computations regardless of which move is used, a known overhead of branchless GPU execution. Verification comprises 2,783 tests across all three levels; 68\% of bugs were caught by L1, 24\% by L2, and 8\% by L3.

\paragraph{HalfCheetah JAX (Gymnasium/MuJoCo $\to$ JAX).} The hardest translation: MuJoCo's HalfCheetah requires articulated-body dynamics (9 DOFs, 7 rigid bodies, 6 actuators) with ground contact. The agent translated forward kinematics, the Composite Rigid Body Algorithm for mass matrices, analytical RNEA for bias forces, and contact Jacobians, all as pure JAX (1k lines, 5 modules). Total translation cost \$3.26 across four solver revisions (penalty-spring, PGS, Jacobi, Newton/LCP), with all 69 tests passing. At matched batch size (32k), our translation achieves throughput parity with MJX ($1.66$M vs.\ $1.6$M SPS) and $5\times$ over Brax at batch 4k. Both our translation and MJX use the same Newton contact solver formulation (acceleration-space QP with Cholesky factorization) and float32 precision; the throughput parity demonstrates that agent-generated, environment-specific code matches the performance of Google's hand-optimized general-purpose engine. The $37\times$ speedup over Gymnasium's single-process CPU execution remains the practically relevant number for training workflows.

\paragraph{TCGJax (Web rules $\to$ Python $\to$ JAX).} TCG Pocket demonstrates specification-to-implementation translation. We extracted rules from official web sources, built a Python reference (29k lines), then translated to JAX (4k lines). The entire translation cost \$4.98 across 11 modules (including an early attempt that erroneously applied rules from a different trading card game; L1 tests and rule verification caught the errors, and additional iterations corrected them). TCG Pocket serves as a contamination control: the Python reference is private (no public repository), so the agent cannot rely on pretraining memorization. The Python reference at 23K SPS (16 processes) is too slow for practical training; JAX at 153K SPS (batch 4K) converges to reward 1.0 in ${\sim}$12 minutes.

\paragraph{Puffer Pong (C $\to$ Rust + JAX).} PufferLib's~\citep{suarez2024pufferlib} C Pong is already optimized (60M SPS random). Translating to JAX enables \texttt{jax.lax.scan}-fused rollouts where the entire rollout compiles into a single GPU kernel with zero CPU$\leftrightarrow$GPU transfer. C environments cannot exploit this fusion. The $42\times$ PPO speedup reflects the CPU-to-GPU architectural change, not like-for-like optimization. This is the core argument for JAX as a target language.

\subsection{Verification Summary}
\label{app:verification_summary}

\begin{table}[ht]
\centering
\caption{\textbf{Verification summary.} L1 = component generation (property tests), L2 = interaction tests, L3 = rollout comparison, Xfer = cross-backend policy transfer.}
\label{tab:verification}
\small
\begin{tabular}{@{}lrrrllll@{}}
\toprule
\textbf{Environment} & \textbf{L1} & \textbf{L2} & \textbf{L3 ep.} & \textbf{Mode} & \textbf{Seeds} & \textbf{Xfer} & \textbf{Status} \\
\midrule
EmuRust        & 32 & 12 & 100 & exact    & 10 & \checkmark & \checkmark \\
PokeJAX        & 1.89k & 670 & 100 & exact & 10 & \checkmark & \checkmark \\
HalfCheetah    & 48 & 12 & 100 & $\epsilon$ ($10^{-3}$) & 10 & \checkmark & \checkmark \\
TCGJax         & 20 & 24 & 100 & exact    & 10 & \checkmark & \checkmark \\
Puffer Pong    & 6 & 3 & 100 & exact    & 10 & \checkmark & \checkmark \\
\bottomrule
\end{tabular}
\end{table}

\subsection{Multi-Agent Validation}
\label{app:multi_agent}

We re-translated Pong with Claude Sonnet~4.6 and HalfCheetah with Claude Opus~4.6, using identical prompts and test suites. Both agents converge to functionally correct translations (Table~\ref{tab:multi_agent}), confirming the methodology is agent-agnostic.

\begin{table}[ht]
\centering
\caption{\textbf{Multi-agent comparison.} Identical inputs; functionally equivalent outputs.}
\label{tab:multi_agent}
\small
\begin{tabular}{@{}llrrr@{}}
\toprule
\textbf{Environment} & \textbf{Agent} & \textbf{Iters} & \textbf{Tests} & \textbf{Cost} \\
\midrule
Pong & Gemini 3 Flash & 13 & 6/6 & \$0.05 \\
     & Claude Sonnet 4.6 & 3 & 5/6$^{\S}$ & ${\sim}$\$0.08 \\
\midrule
HalfCheetah & Gemini 3 Flash & 20 & 69/69 & \$3.26 \\
            & Claude Opus 4.6 & 6 & 69/69 & ---$^{\dagger}$ \\
\bottomrule
\end{tabular}

\vspace{1mm}
{\footnotesize $^{\S}$Same statistical test applied to both agents. $^{\dagger}$Cost not separately tracked.}
\end{table}

\subsection{Verification Ablation Details}
\label{sec:exp:ablation_detail}

\paragraph{HalfCheetah (6-DOF, complex physics).} The L3-only run used 8 end-to-end tests and consumed 42 agent iterations over 35 minutes (\$0.17) without converging. The agent could not isolate dynamics bugs (Coriolis force sign errors, contact Jacobian issues) from end-to-end rollout failures, cycling through vectorization rewrites and stability patches. In contrast, the hierarchical translation converged in 5 iterations (\$0.82, all 69 tests passing), $8.4\times$ faster in iteration count. The L3-only agent's failure mode, code that passes shape and API tests but produces unstable dynamics, is precisely what L1 property tests catch immediately (e.g., mass matrix symmetry, bias force magnitude bounds).

\paragraph{Pong (simple game logic).} The L3-only run converged in 15 iterations over 8.4 minutes (\$0.047). The hierarchical translation converged in 13 iterations over 3.5 minutes (\$0.050) with all 6 tests passing. L3-only succeeded but required 15\% more iterations and $2.4\times$ longer wall-clock time due to reliance on coarse statistical feedback rather than fine-grained L1 signals.

Two data points spanning simple logic (Pong) and moderate physics (HalfCheetah, 6-DOF) consistently show that L3-only fails when contact dynamics and multi-body kinematics chains are involved. The complexity threshold appears to lie between simple game logic and rigid-body physics with $\geq$6 degrees of freedom.

\subsection{Translation Algorithm}
\label{app:algorithm}

Algorithm~\ref{alg:translate} formalizes the closed-loop translation process described in \S\ref{sec:method:verification}.

\begin{algorithm}[ht]
\caption{\textbf{Hierarchical translation and verification.}}
\label{alg:translate}
\begin{algorithmic}[1]
\REQUIRE Reference environment $E_{\text{ref}}$, modules $\{m_1, \dots, m_K\}$ in dependency order, test specifications $\mathcal{T}_1, \mathcal{T}_2, \mathcal{T}_3$, max iterations $T$, episode count $N$
\ENSURE Performance environment $E_{\text{perf}}$ satisfying equivalence

\medskip
\STATE \textbf{Phase 1: Module translation (Level 1)}
\FOR{$k = 1$ \TO $K$}
    \STATE $m_k' \leftarrow \textsc{Agent}(m_k, L_{\text{tgt}})$ \COMMENT{Translate module $m_k$ to target language}
    \FOR{$t = 1$ \TO $T$}
        \IF{$\textsc{RunTests}(\mathcal{T}_1, m_k')$ = \textsc{Pass}}
            \STATE \textbf{break}
        \ELSE
            \STATE $m_k' \leftarrow \textsc{Agent}(\text{failures}, m_k')$ \COMMENT{Repair using L1 diagnostics}
        \ENDIF
    \ENDFOR
    \IF{$t = T$}
        \STATE Request human intervention for module $m_k$
    \ENDIF
\ENDFOR

\medskip
\STATE \textbf{Phase 2: Integration (Level 2)}
\STATE $E_{\text{perf}} \leftarrow \textsc{Compose}(m_1', \dots, m_K')$
\FOR{$t = 1$ \TO $T$}
    \IF{$\textsc{RunTests}(\mathcal{T}_2, E_{\text{perf}})$ = \textsc{Pass}}
        \STATE \textbf{break}
    \ELSE
        \STATE Identify failing module(s); repair while preserving L1 correctness
    \ENDIF
\ENDFOR

\medskip
\STATE \textbf{Phase 3: Validation (Level 3)}
\FOR{$t = 1$ \TO $T$}
    \STATE Run $N$ episodes in $E_{\text{ref}}$ and $E_{\text{perf}}$ with matched seeds and actions
    \IF{all per-step outputs match (exact or within $\epsilon$)}
        \STATE \textbf{break}
    \ELSE
        \STATE Root-cause analysis: add targeted L1/L2 tests; repair and re-verify L1, L2
    \ENDIF
\ENDFOR

\medskip
\STATE \textbf{Phase 4: Cross-backend validation (Level 4)}
\REPEAT
    \STATE Train policy $\pi$ in $E_{\text{perf}}$
    \STATE Evaluate $\pi$ in $E_{\text{ref}}$; compute reward gap $\Delta$
    \IF{$\Delta$ is statistically significant}
        \STATE Diagnose sim-to-sim gap; add targeted L1/L2 tests
        \STATE \textbf{go to} Phase 1 with new tests
    \ENDIF
\UNTIL{$\Delta$ is not statistically significant}
\RETURN $E_{\text{perf}}$
\end{algorithmic}
\end{algorithm}

\subsection{Experimental Details}
\label{app:exp_details}

\emph{Throughput measurement.} All JAX benchmarks exclude one-time JIT compilation from steady-state timing (warm-up call before measurement). JIT compilation ranges from ${\sim}$3\,s (HalfCheetah) to ${\sim}$45\,s (PokeJAX); for a 10-minute training run, this amortizes to ${<}1\%$ for all environments. For PokeJAX (45\,s JIT), amortization over a typical 30-minute run adds ${\sim}2.5\%$.

\emph{GPU memory.} HalfCheetah ${\sim}$4\,GB (65K batch), Pong ${\sim}$2\,GB, PokeJAX ${\sim}$28\,GB (65K), TCGJax ${\sim}$8\,GB (16K).

\emph{Training hyperparameters.} Learning rate $2.5 \times 10^{-4}$, clip ratio 0.2, 4 epochs, GAE $\lambda = 0.95$, $\gamma = 0.99$, with environment-specific batch sizes matched between backends.

\subsection{PufferLib Detailed Comparisons}
\label{app:puffer_detail}

\begin{table}[h]
\centering
\caption{\textbf{PufferLib comparisons.} ``PufferLib training'' reports their full pipeline; matched rows use ${\sim}$2M GRU. All on 1$\times$ RTX 5090.}
\label{tab:puffer}
\small
\begin{tabular}{@{}llrrrr@{}}
\toprule
\textbf{Environment} & \textbf{Benchmark} & \textbf{PufferLib C} & \textbf{Rust} & \textbf{JAX} & \textbf{Speedup} \\
\midrule
\multirow{4}{*}{Puffer Pong}
  & Random (env only) & 60M & 122M & 275M & $4.6\times$ \\
  & PufferLib training & 2.4M (134K) & --- & --- & --- \\
  & GRU Rollout (2M) & 4.5M & 4.5M & 140M & $31\times$ \\
  & GRU PPO (2M) & 854K & 855K & 35.5M & $42\times$ \\
\bottomrule
\end{tabular}
\end{table}

\subsection{Cross-Hardware Validation}
\label{app:hardware}

\begin{table}[h]
\centering
\caption{\textbf{A6000 Ada throughput.} Peak SPS at batch 65k. No code changes required.}
\label{tab:cross_hardware}
\small
\begin{tabular}{@{}lrr@{}}
\toprule
\textbf{Environment} & \textbf{A6000 Ada SPS} & \textbf{vs.\ Reference} \\
\midrule
HalfCheetah JAX & 13.1M & $290\times$ (vs.\ Gymnasium 45K) \\
Pong JAX (scan) & 1.4B & $23k\times$ (vs.\ C 60M env-only) \\
CartPole JAX (scan) & 7.1B & $43k\times$ (vs.\ Gymnasium 164K) \\
\bottomrule
\end{tabular}
\end{table}

\subsection{TCG Pocket Agent Translation Metrics}
\label{app:tcg_metrics}

The TCG Pocket translation was conducted entirely through logged sessions with Gemini~3 Flash Preview. \emph{Phase~1} translated five core modules (1.4k source lines) via programmatic API calls: 20 iterations consuming 83K tokens (\$0.02). \emph{Phase~2} used the Gemini CLI to translate six logic-heavy modules: 29.3M input tokens across 256 messages (\$4.96), with 79--95\% cache hit rates.

\begin{figure}[h]
\centering
\includegraphics[width=\textwidth]{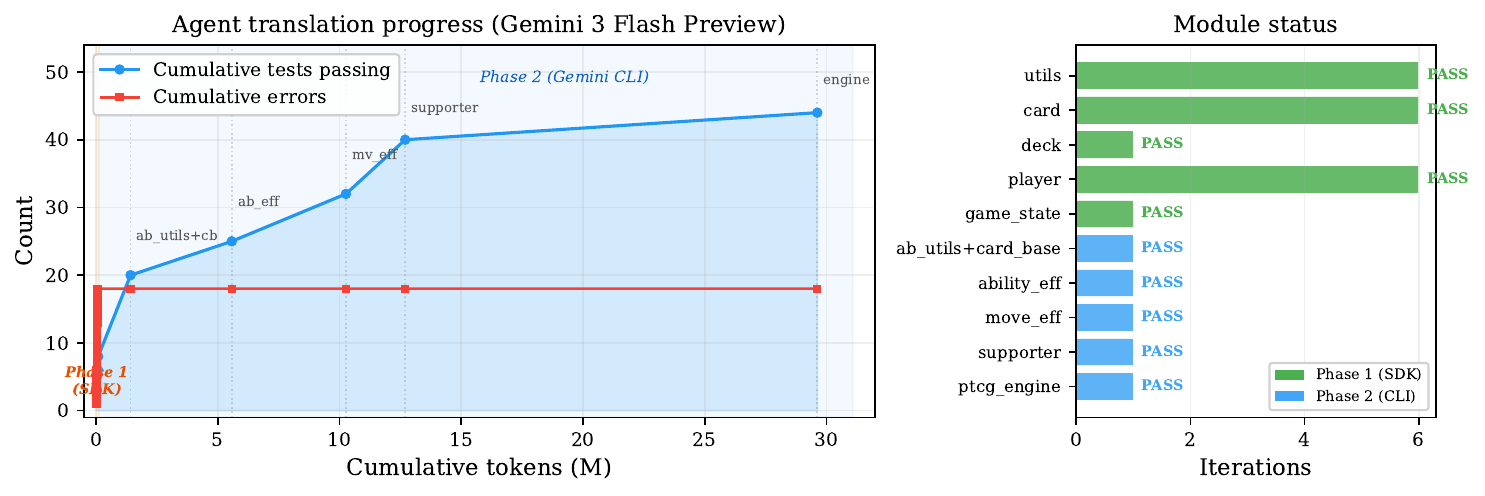}
\caption{\textbf{Agent translation metrics for TCG Pocket.} Cumulative L1 tests passing vs.\ tokens consumed. Total: \$4.98 for 4k lines.}
\label{fig:agent_metrics}
\end{figure}

\subsection{Test Coverage}
\label{app:coverage}
\begin{table}[h]
\centering
\caption{\textbf{Test coverage by environment.} Pass/Total counts all \texttt{pytest}-collected test functions, including parametrized variants and regression tests added after translation (hence larger than Table~\ref{tab:effort}'s translation-time counts). Line coverage measured with \texttt{pytest-cov} (Python) or \texttt{cargo test} (Rust). $^*$PokeJAX: JIT-compiled dispatch prevents line-level instrumentation; figure reflects only the directly instrumented mechanics/core modules. $^\dagger$EmuRustGBA: 110 unit tests in Rust source; 85 integration tests and 23 hardware-feature tests exercise the PyO3 Python bindings. $^{\ddagger}$Failing tests: Pong's 1 failure is a statistical distribution test sensitive to sample size (same test fails for both Gemini and Claude translations, Table~\ref{tab:multi_agent}); HalfCheetah's 4 failures are tight-tolerance parametrized tests affected by float32 vs.\ float64 differences (all core L1/L2/L3 tests pass).}
\label{tab:coverage}
\small
\begin{tabular}{@{}lrrrl@{}}
\toprule
\textbf{Environment} & \textbf{Pass/Total} & \textbf{Stmts} & \textbf{Coverage} & \textbf{Notes} \\
\midrule
CartPole JAX     & 9/9     & 107   & 60\%  & main(), rendering untested \\
Pong JAX         & 5/6     & 158   & 73\%  & 1 statistical test$^{\ddagger}$ \\
HalfCheetah JAX  & 136/140 & 462   & 96\%  & 4 float32 tolerance$^{\ddagger}$ \\
TCG Pocket JAX   & 50/50   & 1k & 77\% & L1/L2/L3 all passing \\
PokeJAX$^*$      & 90/95   & 1k & 53\% & JIT limits instrumentation \\
EmuRustGBA$^\dagger$ & 218/218 & 527 & 98\% & 110 Rust + 108 Python tests \\
\midrule
\textbf{Total}   & \textbf{508/518} & & & \\
\bottomrule
\end{tabular}
\end{table}

\section{Performance Optimization Guide}
\label{app:optimization}

After a translated environment passes all three verification levels, the next step is performance optimization. This appendix provides concrete techniques and a reusable agent prompt for maximizing environment throughput. The techniques are organized by target backend: JAX (GPU) and Rust (CPU).

\subsection{JAX Optimization Checklist}
\label{app:optimization:jax}

The following patterns, distilled from our case studies, consistently improve JAX environment throughput. They are ordered by typical impact.

\paragraph{1. Fixed-size state arrays.} JAX requires array shapes to be known at compile time. Replace all dynamic-length data structures (lists, dicts with varying keys, variable-length arrays) with fixed-size \texttt{jnp.ndarray} fields padded to maximum capacity. Use a sentinel value (e.g., \texttt{-1} or \texttt{NO\_CARD\_ID}) for unused slots. In TCG Pocket, this reduced card zone storage from Python lists to fixed \texttt{(MAX\_HAND\_SIZE,)} arrays, enabling JIT compilation of the entire game engine.

\paragraph{2. Branchless conditionals with \texttt{jnp.where}.} Replace Python \texttt{if}/\texttt{else} with \texttt{jnp.where(condition, true\_val, false\_val)}. Both branches are computed and the result is selected by mask; this is faster on GPU because it avoids warp divergence. For multi-way branches, use nested \texttt{jnp.where} or \texttt{jax.lax.switch}. Reserve \texttt{jax.lax.cond} for unbatched cases where one branch is significantly more expensive (it evaluates only the selected branch). Note that under \texttt{vmap}, \texttt{lax.cond} evaluates both branches regardless, because different batch elements may take different paths; in batched contexts, \texttt{jnp.where} is preferred. In Puffer Pong, all ball-paddle collision logic uses \texttt{jnp.where}:
\begin{verbatim}
  ball_vy = jnp.where(wall_hit, -ball_vy, ball_vy)
  ball_vx = jnp.where(paddle_hit, -ball_vx, ball_vx)
\end{verbatim}

\paragraph{3. \texttt{vmap} for batch parallelism.} Write environment logic for a \emph{single} instance, then apply \texttt{jax.vmap} to vectorize across the batch dimension. This generates fused GPU kernels that process all environments in one call. Mark shared constants (terrain maps, card databases) with \texttt{in\_axes=None} so they are broadcast rather than duplicated:
\begin{verbatim}
  step_batch = jax.vmap(step_single, in_axes=(0, 0))
  step_with_terrain = jax.vmap(
      partial(step, terrain=terrain),
      in_axes=(0, 0)  # terrain not batched
  )
\end{verbatim}

\paragraph{4. JIT the outer interface.} Apply \texttt{jax.jit} to the \texttt{vmapped} step and reset functions so the entire batch operation compiles to a single GPU kernel. Pre-compile during initialization to avoid first-call latency during training:
\begin{verbatim}
  self._step_jit = jax.jit(step_batch)
  self._reset_jit = jax.jit(reset_batch)
  # Warmup: call once with dummy data
  _ = self._step_jit(dummy_states, dummy_actions)
\end{verbatim}

\paragraph{5. \texttt{lax.scan} for multi-step fusion.} When the training loop calls \texttt{env.step} inside a rollout loop, fuse the loop with \texttt{jax.lax.scan} to compile the entire rollout into one kernel. This eliminates per-step CPU$\to$GPU dispatch overhead. In CartPole, this improved throughput by $3.2\times$ over a Python loop calling \texttt{jit}ted steps:
\begin{verbatim}
  def scan_body(states, actions_t):
      states, rewards, terminals = step_batch(states, actions_t)
      return states, (rewards, terminals)
  rollout = jax.jit(partial(jax.lax.scan, scan_body))
\end{verbatim}

\paragraph{6. Minimize data types.} Use \texttt{int8} for categorical state (entity types, directions, flags) and \texttt{float32} only for values requiring arithmetic. For example, using \texttt{int8} for categorical entity fields can reduce per-environment state significantly, improving memory bandwidth utilization.

\paragraph{7. Pre-allocate reward and observation buffers.} Initialize all output arrays (rewards, terminals, observations) as zeros in the state. Update in-place with \texttt{.at[].set()} rather than creating new arrays. Avoid \texttt{jnp.concatenate} or \texttt{jnp.stack} in the hot path.

\paragraph{8. Normalize observations at the source.} Compute normalized observations inside the JIT-compiled step function rather than in a separate Python post-processing step. Pre-compute constant denominators:
\begin{verbatim}
  PADDLE_RANGE = MAX_PADDLE_Y - MIN_PADDLE_Y  # constant
  obs_paddle = (state.paddle_y - MIN_PADDLE_Y) / PADDLE_RANGE
\end{verbatim}

\subsection{Rust Optimization Checklist}
\label{app:optimization:rust}

\paragraph{1. Rayon \texttt{par\_iter} for environment parallelism.} Use \texttt{rayon::prelude::par\_iter\_mut} to step all environments in parallel across CPU cores. Each environment is independent, making this embarrassingly parallel:
\begin{verbatim}
  self.emulators.par_iter_mut()
      .zip(actions.iter())
      .for_each(|(emu, &action)| emu.step(action));
\end{verbatim}
This typically provides near-linear scaling up to the number of physical cores ($8{-}16\times$).

\paragraph{2. Pre-allocate observation buffers.} Allocate observation, reward, and terminal buffers once at initialization, then reuse every step via slice copies. Avoid \texttt{Vec::push} or allocation in the step loop:
\begin{verbatim}
  let obs_buffer = vec![0u8; num_envs * OBS_SIZE];
  // In step(): copy directly into pre-allocated slice
  obs_buffer[i*OBS_SIZE..(i+1)*OBS_SIZE]
      .copy_from_slice(&emu.get_obs());
\end{verbatim}

\paragraph{3. Frame skip without rendering.} For emulator environments, implement a fast path that skips PPU/rendering for intermediate frames. Only render the final frame that produces the observation. In EmuRust, this saved ${\sim}60\%$ of per-step time at frame skip 24:
\begin{verbatim}
  emu.run_frames_no_render(frame_skip - 1); // fast path
  emu.run_frame();                          // render last frame
\end{verbatim}

\paragraph{4. Lookup tables for game mechanics.} Replace computed game logic with pre-computed \texttt{const} arrays. For example, element-type effectiveness matrices, passability checks, and noise gradients can all be pre-computed as static lookup tables:
\begin{verbatim}
  const EFFECT_MATRIX: [[i32; 5]; 5] = [[1,1,1,1,1], ...];
  let damage_mult = EFFECT_MATRIX[atk_type][def_type];
\end{verbatim}

\paragraph{5. \texttt{\#[inline(always)]} on hot functions.} Mark observation writing, single-step physics, and reward computation as \texttt{\#[inline(always)]} to eliminate function call overhead in tight loops. Profile first; only inline functions called millions of times per second.

\paragraph{6. \texttt{Arc<Vec<>>} for shared immutable data.} When each environment instance needs access to large immutable data (ROM images, card databases, terrain maps), wrap it in \texttt{Arc} and clone the reference:
\begin{verbatim}
  let rom = Arc::new(rom_data);
  let emulators: Vec<_> = (0..num_envs)
      .map(|_| Emulator::new(rom.clone()))
      .collect();
\end{verbatim}
One copy in memory regardless of batch size.

\paragraph{7. Compact struct layout.} Separate hot data (accessed every step) from cold data (accessed occasionally). Keep entity structs small; use \texttt{i32} instead of \texttt{i64}, pack booleans into bitfields or \texttt{i32} flags. This improves L1/L2 cache utilization.

\paragraph{8. Efficient PyO3 bindings.} For the Python$\leftrightarrow$Rust boundary: accept NumPy arrays via \texttt{PyReadonlyArrayN} (zero-copy read), return observations by writing directly into a pre-allocated NumPy array via \texttt{PyArrayN::as\_slice\_mut()}. Minimize the number of Python$\to$Rust calls per step (one call for all environments, not one per environment).

\subsection{Optimization Agent Prompt}
\label{app:prompts:optimize}

The following prompt is used after the environment passes Level~1--3 verification. It instructs the coding agent to optimize throughput without changing the simulator.

\begin{quote}
\small\ttfamily
The [JAX/Rust] environment implementation has passed all verification tests (Level~1 property tests, Level~2 interaction tests, Level~3 rollout comparison). Now optimize it for maximum steps-per-second (SPS) throughput.

\textbf{Current performance:} [X] SPS at batch size [B] on [hardware].

\textbf{Target:} Maximize SPS while maintaining all existing tests passing.

\textbf{Constraints:}
\begin{itemize}
\item[-] All Level~1, 2, and 3 tests must continue to pass after optimization
\item[-] Do not change the environment's external API (step, reset, observation/reward shapes)
\item[-] Do not change game simulator or reward logic
\end{itemize}

\textbf{[For JAX environments] Apply these optimizations in order:}
\begin{enumerate}
\item Replace any remaining Python \texttt{if}/\texttt{else} on JAX values with \texttt{jnp.where} or \texttt{jax.lax.cond}
\item Ensure all state arrays have static shapes (no dynamic allocation)
\item Apply \texttt{jax.vmap} for batch parallelism over a single-instance step function
\item Wrap the vmapped function with \texttt{jax.jit}
\item Reduce data types: use \texttt{int8} for categorical fields, \texttt{float32} only for arithmetic
\item Pre-compute observation normalization constants
\item Profile with \texttt{jax.profiler} and eliminate remaining bottlenecks
\end{enumerate}

\textbf{[For Rust environments] Apply these optimizations in order:}
\begin{enumerate}
\item Add \texttt{rayon} dependency and parallelize \texttt{step}/\texttt{reset} with \texttt{par\_iter\_mut}
\item Pre-allocate all output buffers (obs, rewards, terminals) at initialization
\item Add \texttt{\#[inline(always)]} to step, observation, and reward functions
\item Replace computed game logic with \texttt{const} lookup tables where applicable
\item Implement frame-skip fast path (skip rendering for intermediate frames)
\item Use \texttt{Arc<Vec<>>} for shared immutable data across environments
\item Profile with \texttt{cargo flamegraph} and eliminate remaining bottlenecks
\end{enumerate}

\textbf{After each optimization:}
\begin{enumerate}
\item Run the full test suite to verify correctness
\item Measure SPS at batch sizes [32, 128, 512, 2048, 8192]
\item Report the speedup from each change
\end{enumerate}

Begin with a profiling analysis to identify the current bottleneck, then apply optimizations targeting that bottleneck first.
\end{quote}

\section{Representative Agent Prompt}
\label{app:prompts}

This appendix presents a representative prompt used during agent-assisted translation (\S\ref{sec:method:verification}). Each prompt follows a \emph{generic template structure} that stays constant across all environments: (1)~source module specification with line count, (2)~target language constraints, (3)~interface contract (function signatures and return types), (4)~reference behavior (source code pasted verbatim), and (5)~instruction to generate Level~1 property tests. The parts that vary per-environment are the module source code, target constraints, and interface contracts, filled in by the human for each module. Analogous templates instantiate the Level~1/2 test-generation and bug-repair prompts; we omit them for brevity. The example below is instantiated for EmuRust.

\subsection{Module Translation Prompt}
\label{app:prompts:translate}

The following prompt initiates translation of a single module. The agent receives the source code, target language constraints, and the module's interface contract.

\begin{quote}
\small\ttfamily
Translate the following Game Boy CPU module from C/Python (PyBoy) to Rust.

\textbf{Source module:} cpu.py (161 lines), SM83 instruction set implementation.

\textbf{Target constraints:}
\begin{itemize}
\item[-] Pure Rust, no unsafe except for FFI boundaries
\item[-] All registers as a struct with public fields (for save/load state)
\item[-] Instruction dispatch via match on opcode byte
\item[-] Return cycle count from each instruction for PPU synchronization
\end{itemize}

\textbf{Interface contract:}
\begin{itemize}
\item[-] \texttt{fn step(\&mut self, mem: \&mut Memory) -> u32}: execute one instruction, return T-cycles
\item[-] \texttt{fn handle\_interrupts(\&mut self, mem: \&mut Memory)}: check and dispatch IF/IE
\end{itemize}

\textbf{Reference behavior (from PyBoy source):}\\
\texttt{[Source code of cpu.py pasted here, 161 lines]}

Begin translation. After completing, write Level~1 property tests that verify each instruction against reference input/output pairs.
\end{quote}

\end{document}